Exploring Generative Adversarial Networks for Image-to-Image Translation in STEM Simulation

Nick Lawrence[2], Mingren Shen[1], Ruiqi Yin[3], Cloris Feng[4], Dane Morgan[1]

[1] Department of Materials Science and Engineering, University of Wisconsin-Madison, Madison, Wisconsin, 53706, USA

[2] Department of Industrial & Systems Engineering, University of Wisconsin-Madison, Madison, Wisconsin, 53706, USA

[3] Department of Computer Science, University of Wisconsin-Madison, Madison, Wisconsin, 53706, USA

[4] Department of Electrical & Computer Engineering, University of Wisconsin-Madison, Madison, Wisconsin, 53706, USA

## Abstract

The use of accurate scanning transmission electron microscopy (STEM) image simulation methods require large computation times that can make their use infeasible for the simulation of many images. Other simulation methods based on linear imaging models, such as the convolution method, are much faster but are too inaccurate to be used in application. In this paper, we explore deep learning models that attempt to translate a STEM image produced by the convolution method to a prediction of the high accuracy multislice image. We then compare our results to those of regression methods. We find that using the deep learning model Generative Adversarial Network (GAN) provides us with the best results and performs at a similar accuracy level to previous regression models on the same dataset. Codes and data for this project can be found in this GitHub repository, https://github.com/uw-cmg/GAN-STEM-Conv2MultiSlice.

# Introduction

As the uses of scanning transmission electron microscopy (STEM) in applications of high-resolution materials imaging continue to expand, methods for simulating STEM images become increasingly important. Simulated images are often used to verify or provide quantitative interpretations for experimental STEM results in areas such as high precision two-dimensional measurements [1], atomic electron tomography [2], and three-dimensional imaging of point defects [3].

STEM images can be simulated in several ways. The multislice method is most commonly used because it is the most accurate simulation method [4]. However, the multislice method requires significant computation time, taking weeks of central processing unit (CPU) time to produce a single image [5]. Efforts have been made to reduce this time, such as fast Fourier transforms and CPU parallelization and have resulted in a 3-fold reduction in CPU time [5,6]. Other methods, such as the convolution method, make assumptions that simplify the physics in the STEM simulation, deciding to trade accuracy for a fast computation time [4].

Several efforts have been made to develop methods that take a convolution image and output an image that is closer to the corresponding multislice image. Yu et al. investigated high-order polynomial models in a one-to-one mapping of pixels from the convolution image to the multislice [7]. Combs et al. generalized this approach by including many-to-one pixel mappings [8]. The model in her work is based on a multivariate polynomial fit, which is simple to understand and extremely rapid to fit and apply. Our work aims to extend that of Combs by instead using machine learning techniques in place of polynomial regression.

Deep learning typically refers to a special area of machine learning that uses neural networks with many layers of nodes to build a complex network that is able to extract or learns the underlining pattern or mapping between input data and output results [9]. Deep learning has shown success in many areas e.g.

automatic driving, speech, medical images[10,11], material sciences[12–16] and image recognition [9] and it even surpasses human performance in some tasks like Go [17]. Generative Adversarial Networks (GANs) are a specific kind of deep learning architecture that provides a unique way to let the neural network learns to generate new data with the same statistics as the training data [18] and has been applied in different areas e.g. image style transfer [19], medical image analysis [20,21], data augmentation [22], and others [23]. In the present work we explore using a GAN to generate the best guess at a high-fidelty multislice image from the corresponding low-fidelity convolution. The GAN employs two neural networks that effectively train each other. The first, coined the 'generator', takes convolution images as input and outputs its prediction of the corresponding multislice image. The second network, coined the 'discriminator', takes in an image that is either a true multislice image or a 'fake' multislice image that was generated by the generator. The discriminator outputs its guess of if the image is real. As the discriminator trains, it will get better at telling the generated images from the real images. So in our project of image-to-image translation of the convolution image, this forces the generator to get better at making 'believable' multislice images.

# Methods

### Data

In order to develop and assess the GAN-basd approach, we generated corresponding convolution and multislice simulated images for sets of Pt nanoparticles, Pt-Mo nanoparticles with ~5% Mo, and Pt-Mo nanoparticles with ~50% Mo. Each set contained nanoparticles of similar size but varying structure, ranging from purely amorphous to purely crystalline. Multislice images of each particle were generated using code from E.J. Kirkland [4], then preprocessed the data so that they are saved in .png files of constant size, 256x256 pixels. Each pixel's "brightness" corresponds to the measure of intensity in the pixel. Intensity is the percent of the total electrons present in the pixel. This data is the same as used in Ref. [8].

## Performance Metrics

We will compare the performance of different deep learning models and compare their results with that of Combs's polynomial model. Combs uses fractional RMSE, that is, RMSE divided by the standard deviation of pixel intensity in the image, to compare the predictions produced by their polynomial fit model to the desired multislice images.

$$\sqrt{\frac{\sum_{i=1}^{N}(predicted_i - actual_i)^2}{N}} \Big/ \sigma$$

*Equation 1: Fractional RMSE*

The benefit of dividing the RMSE by the standard deviation is that the metric provides an estimate of error relative to the variation in the sample, which will tend to produce a comparable value for similar quality models across samples with widely varying intensity scales and variability. Our pixel intensity values lie in the range of [0, .01] while traditional image values lie in the range of [0, 255]. Thus, we must normalize the errors so that our metric is comparable to traditional images as well as the previous work of Combs. We present these errors as a % error, i.e., we multiple Equation 1 by 100x.

We also considered a metric called Structural Similarity Index (SSIM) that uses a more complex formula to quantify the visual similarity of two images and which reflects the visual similarity between two images better than mean squared error [24].

## Model

In addition to assessing again the model from Combs with additional metrics, in this work we have focused on a generative adversarial network (GAN) method. Our GAN is based off the work of Erik

Lindernoren (https://github.com/eriklindernoren/Keras-GAN), who programmed a GAN based off the work of Isola et al, who adapted the original GAN to suit image-to-image translation problems [25].

## Results and Discussion

To evaluate the accuracy of our model, we performed a k-fold cross validation over 5 folds. For each fold we train the GAN on the training data and predict the validation data and calculate the error from Equation 1. Table 1 shows the GAN's performance on the k-fold cross validation.

| Fold | Frac RMSE |
|---|---|
| 1 | 7% |
| 2 | 8% |
| 3 | 12% |
| 4 | 9% |
| 5 | 18% |

*Table 1: GAN performance on 5-fold cross validation*

Although the fold contained a random assortment of test images, the performance in the folds varied between 7% and 18%, for an average performance of 10.8%. Comparing this to Combs' results, our model performs similarly to Combs' optimized model, who averaged slightly over 9% error [8]. We did not perform a full k-fold cross validation for the SSIM metric, although we found that the SSIM when using first fold as the test set was 0.89, a quite good value (see Figure 1).

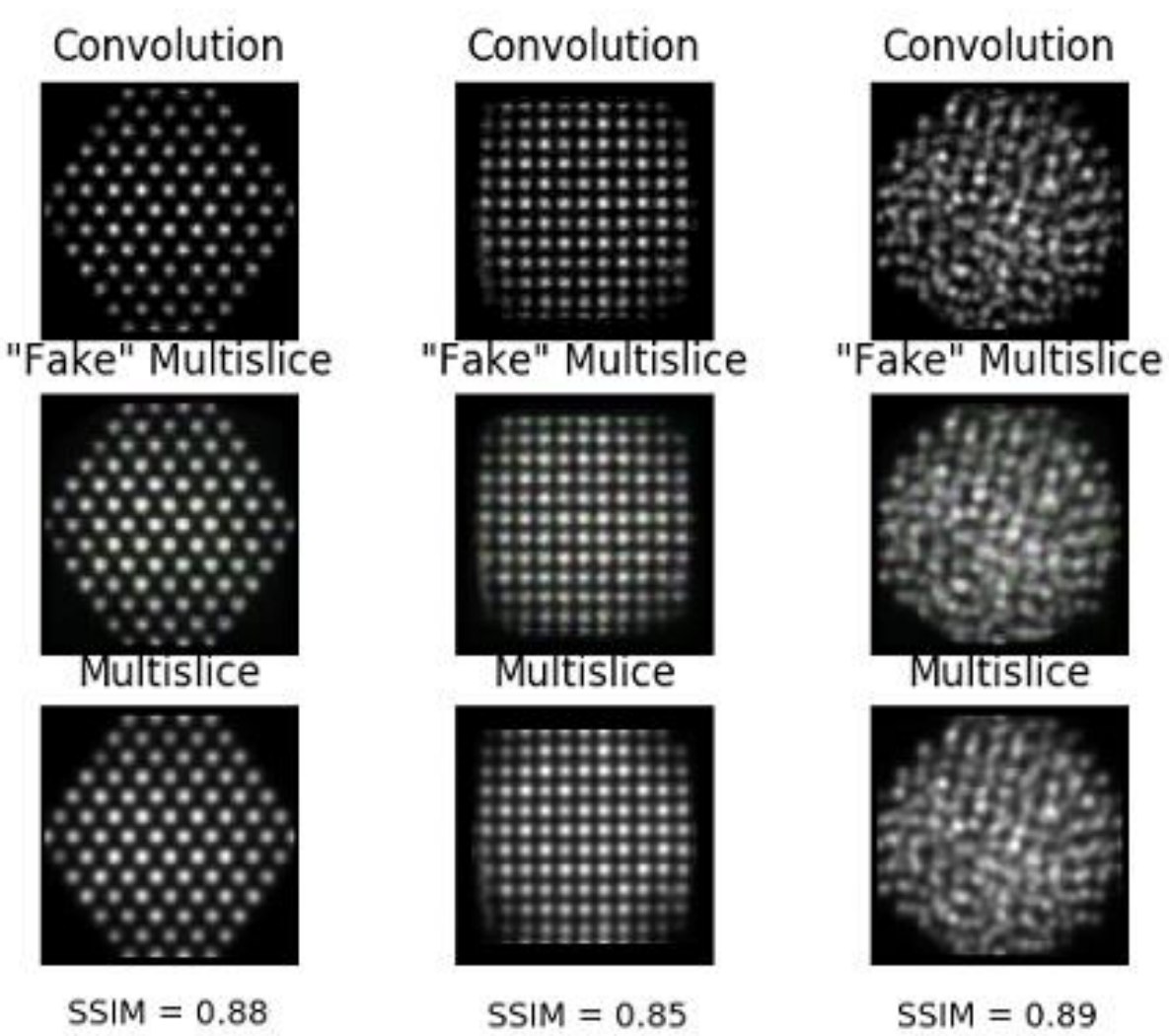

*Figure 1: SSIM evaluations of 3 test images*

Before performing the k-fold cross validation, we performed a modest level of hyperparameter optimization where we chose the hyperparameters that optimized the errors from Equation 1 when using the first fold as the test data set. We performed a grid search hyperparameter optimization for four different hyperparameters listed in Table 2:

| **Hyperparameter** | **Values Tested** | **Optimal Parameter** |
|---|---|---|
| Learning Rate | 0.002, 0.0002, 0.00002 | 0.0002 |
| Batch Size | 1, 2, 3, 5, 8, 10, 12, 13, 15, 20 | 3 |
| Optimizer | Adam, Stochastic Gradient Descent | Adam |
| Learning Metric | MAE, MSE, MAE 50:50 MSE | MSE |

*Table 2: Hyperparameter Optimization Parameters*

Through our testing, we found that batch size made the most significant difference on model performance. Not only did optimizing batch size decrease the error of the model, but it also improved the consistency of the performance of the model at each epoch. We qualitatively assessed the consistency of the model by graphing the error of the model's predictions of the test set after each epoch, i.e. a set of iterations of training. As seen in Figure 2, at the optimal batch size of 3, the error on the test set is more stable as the epochs increase than for batch sizes of 1 or 8. At this optimal batch size the performance of the model will be less dependent on the exact number of epochs trained than for other batch sizes. Our approach to hyperparameter optimization technically allows for data leakage as it changes the approach knowing scores on all the data. This might lead to some overestimation of our performance, although similar optimization approaches and data leakage concerns were present in the work by Combs et al. in Ref. [8] so our approach is reasonable given the goal of comparing to that work.

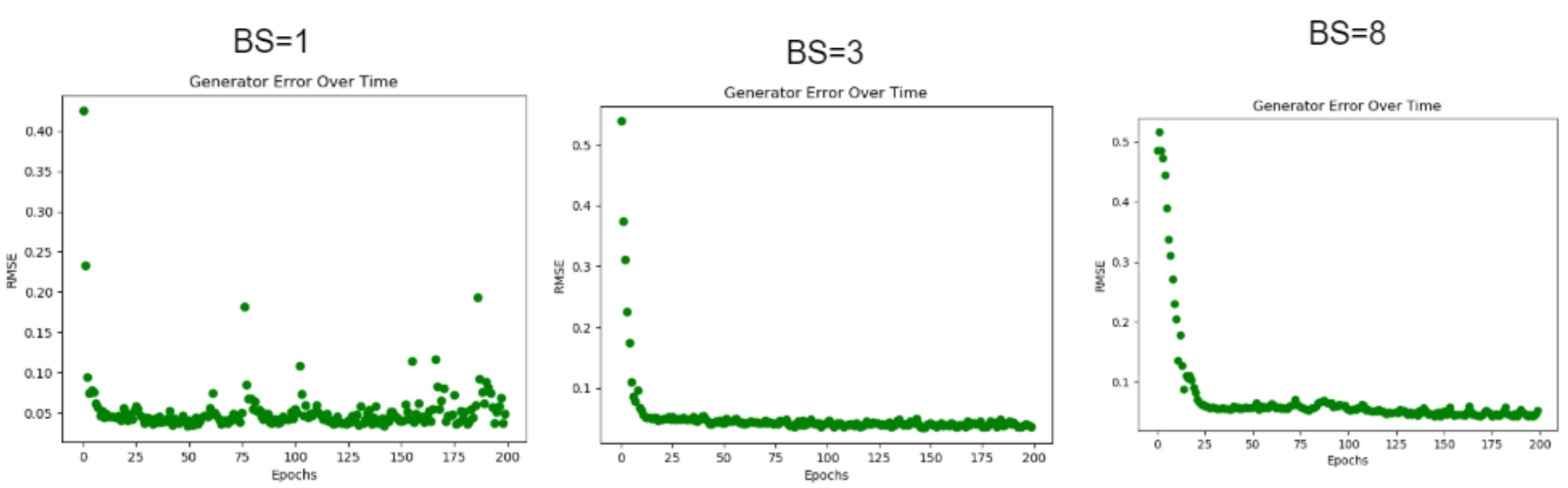

*Figure 2: GAN performance with different batch sizes.*

## Conclusion

Deep learning techniques are revolutionizing the world of predictive modeling. In their short lifetime, they have performed better than simpler regression techniques in many cases. In this work we explored using deep learning to learn accurate multislice STEM simulations from less accurate convolution STEM simulations. we found that a Generative Adversarial model performs as well as a previous simpler regression model on the same dataset. The GAN model also produced predictions of the multislice images that had extremely high visual similarity scores with the true multislice simulations. We would recommend any group working on a similar problem to consider testing a GAN along with more traditional regression methods on their dataset to see which performs best.

## Acknowledgements

We are very grateful to Prof. Paul Voyles at University of Wisconsin for his extensive help with this work.